\documentclass[conference]{IEEEtran}
\usepackage{amssymb,amsmath}
\usepackage{cite}
\usepackage{graphicx}
\usepackage[caption=false]{subfig}
\usepackage[font=footnotesize]{caption}
\usepackage{psfrag}
\usepackage{url}
\usepackage[latin1]{inputenc}
\usepackage[absolute,overlay]{textpos}
\usepackage{tikz}
\usetikzlibrary{arrows,calc,decorations.markings}
\usetikzlibrary{topaths}
\usetikzlibrary{shapes,trees}
\usetikzlibrary{arrows}
\usetikzlibrary{shadows}
\usetikzlibrary{positioning}
\usetikzlibrary{matrix}
\usetikzlibrary{shapes.geometric}
\usetikzlibrary{decorations.pathmorphing}
\usepgflibrary{patterns}
\usetikzlibrary{calc}
\usetikzlibrary{fit}					% fitting shapes to coordinates

\DeclareMathOperator*{\argmax}{arg\,max}
\usepackage{pgf}
\usetikzlibrary{arrows,automata}
\usepackage[latin1]{inputenc}

\usepackage{textcomp}
\usepackage{xcolor}
\def\BibTeX{{\rm B\kern-.05em{\sc i\kern-.025em b}\kern-.08em
    T\kern-.1667em\lower.7ex\hbox{E}\kern-.125emX}}

\usepackage{amsthm}

\usepackage{tabularx}
\usepackage{makecell}
\usepackage{multirow}

\usepackage{amsmath}
\usepackage{amssymb}
\usepackage{bbm}
\usepackage{bm}
\usepackage{comment}
\usepackage[capitalize]{cleveref}

\usepackage{pgfplots}
\usepackage{float}
\usepackage{booktabs}

\pgfplotsset{compat=1.15}

\usepackage[bottom=1.1 in,top=0.75in,left=0.63in,right=0.63 in]{geometry}

\DeclareMathOperator{\Tr}{Tr}

\begin{document}

\title{Age Minimization in Massive IoT via UAV Swarm: A Multi-agent Reinforcement Learning Approach}
\author{
	\IEEEauthorblockN{Eslam Eldeeb, Mohammad Shehab and Hirley Alves \\
	}
	\IEEEauthorblockA{Centre for Wireless Communications (CWC), University of Oulu, Finland \\
	Email: firstname.lastname@oulu.fi}
}
\maketitle

\begin{abstract}
In many massive IoT communication scenarios, the IoT devices require coverage from dynamic units that can move close to the IoT devices and reduce the uplink energy consumption. A robust solution is to deploy a large number of UAVs (UAV swarm) to provide coverage and a better line of sight (LoS) for the IoT network. However, the study of these massive IoT scenarios with a massive number of serving units leads to high dimensional problems with high complexity. In this paper, we apply multi-agent deep reinforcement learning to address the high-dimensional problem that results from deploying a swarm of UAVs to collect fresh information from IoT devices. The target is to minimize the overall age of information in the IoT network. The results reveal that both cooperative and partially cooperative multi-agent deep reinforcement learning approaches are able to outperform the high-complexity centralized deep reinforcement learning approach, which stands helpless in large-scale networks.
\end{abstract}
\begin{IEEEkeywords}
	Age of information, UAVs, Machine learning, Multi-agent reinforcement learning
\end{IEEEkeywords}

\section{Introduction}\label{sec:introduction}
The road to future wireless networks is encompassed by a massive deployment of IoT devices to enable vehicle platooning, smart agriculture, and yet many unforeseen applications. In such applications, there is an imperative requirement for fresh information about processes monitored or executed by these devices. Hence, information freshness has been the focus of many studies in the recent few years \cite{petar_perspective,AoI}. In this context, a metric termed Age of Information (AoI) was introduced to address time sensitivity in different scenarios. AoI is defined as the time elapsed since the generation of the freshest received packet. Thus, the lower the age, the fresher the information available about a certain process.

Meanwhile, the deployment of UAVs as mobile relay units or base stations (BS) is considered to be appealing for remote area coverage and collecting data from low-energy devices due to the following reasons \cite{UAVs}:
\begin{enumerate}

    \item UAVs can provide coverage for remote areas, where it is cumbersome to replace the batteries of IoT devices.

    \item UAVs can reduce the transmission distance of IoT nodes by moving close to them and then relaying the transmitted information to the BS. 

    \item UAVs can reach high altitudes and hence, the probability of line-of-sight (LOS) with both BS and IoT nodes becomes higher. 
    
\end{enumerate}
In this context, UAV swarming is also suggested as a promising coverage solution for massive MIMO \cite{Swarm_MIMO},  edge intelligence \cite{AoI_swarm}, as well as UAV search and rescue operations \cite{swarm_search_rescue}.

\subsection{Literature Review}
\vspace{-0mm}
Many works have studied age minimization in UAV-assisted IoT. For instance, the work in \cite{multi-uav_age} suggested a graph theory approach to minimize the age, where the UAVs collect data from sensors only at fixed data collection points around the map, without the flexibility of collecting information anywhere. Similarly, the work in \cite{UAVs_stationary} optimized the stationary positions of these UAVs using a game theoretic approach. Meanwhile, the authors of \cite{AoI_genetic} applied the genetic algorithm and dynamic programming to address the same problem. However, their approach was applied to a limited number of only 15 sensor nodes with no scalability to massive deployments. The same shortcoming also applies to the work in \cite{deep_china}.

To this end, the massive deployment of devices and services is overwhelmingly leading to large-scale problems with a large number of non-linear parameters, making them computationally prohibitive for optimization using conventional statistical methods. Such methods are increasingly becoming unable to scale well to these problems, which mandates a paradigm shift towards more scalable approaches. At the heart of these approaches lie the decentralized machine learning algorithms. Decentralized approaches such as Federated learning and multi-agent reinforcement learning (MARL) usually have less complexity \cite{DL}. Furthermore, decentralized solutions may not require sharing of all information about agents, and hence, data privacy is preserved. In the recent few years, MARL has drawn a great amount of attention to solving problems related to massive IoT. For example, the authors \cite{energy_harvesting} applied MARL to propose an energy harvesting scheme in massive scenarios. In addition, the work in \cite{AoI_swarm2} applied deep MARL in a UAV swarm sensing application.

\vspace{0 mm}
\subsection{Paper Contributions}
We summarize the contributions of this work as follows\footnote{Unlike our previous works in~\cite{9815722,9950310}, which present two UAVs serve $10$ IoT devices and a centralized solution for multiple UAVs serving a cluster-based IoT network, respectively, we focus on comparing MARL solutions in massive IoT network served by a swarm of UAVs.}:
\begin{itemize}
    \item We propose multi-agent deep reinforcement learning solutions with partial and no sharing of information to minimize the AoI.
    
    \item Our solutions account for the IoT devices' energy consumption, UAV transmission modes, as well as the division of the massive network into clusters.
    \item We compare the proposed MARL approaches to the centralized solution in terms of performance and complexity.
    \item Interestingly, the proposed MARL approaches render very good performance in terms of AoI and complexity, when compared to centralized RL, especially when deploying a large number of UAVs.
\end{itemize}
\subsection{Outline}
Section~\ref{sec:sysmodel} depicts the system model and the problem formulation. The proposed deep MARL schemes are presented in Section \ref{sec:DRL}. Section \ref{results} elucidates the results and finally, the paper is concluded in Section~\ref{conclusions}.
\vspace{0 mm}
%Deep RL can be applied to reduce the state space in extremely high dimensional problems. A typical example is the large-scale UAV trajectory planning problem for serving massive MTC deployment with URLLC requirements, which has been recently addressed in works such as \cite{CNN}, where the authors applied combinatorial neural networks in order to optimize the user scheduling and AoI, and \cite{Eldeeb2022}, where deep RL and device clustering are used to facilitate trajectory planning for multiple UAVs with the target of reducing the AoI and energy consumption of MTC networks. These works imposed maximum thresholds for AoI to guarantee minimum level of reliability.
\vspace{0 mm}
\section{System model and Problem Formulation}\label{sec:sysmodel}
\vspace{0 mm}
\begin{figure}[t!]
    \centering    \includegraphics[width=0.80\columnwidth]{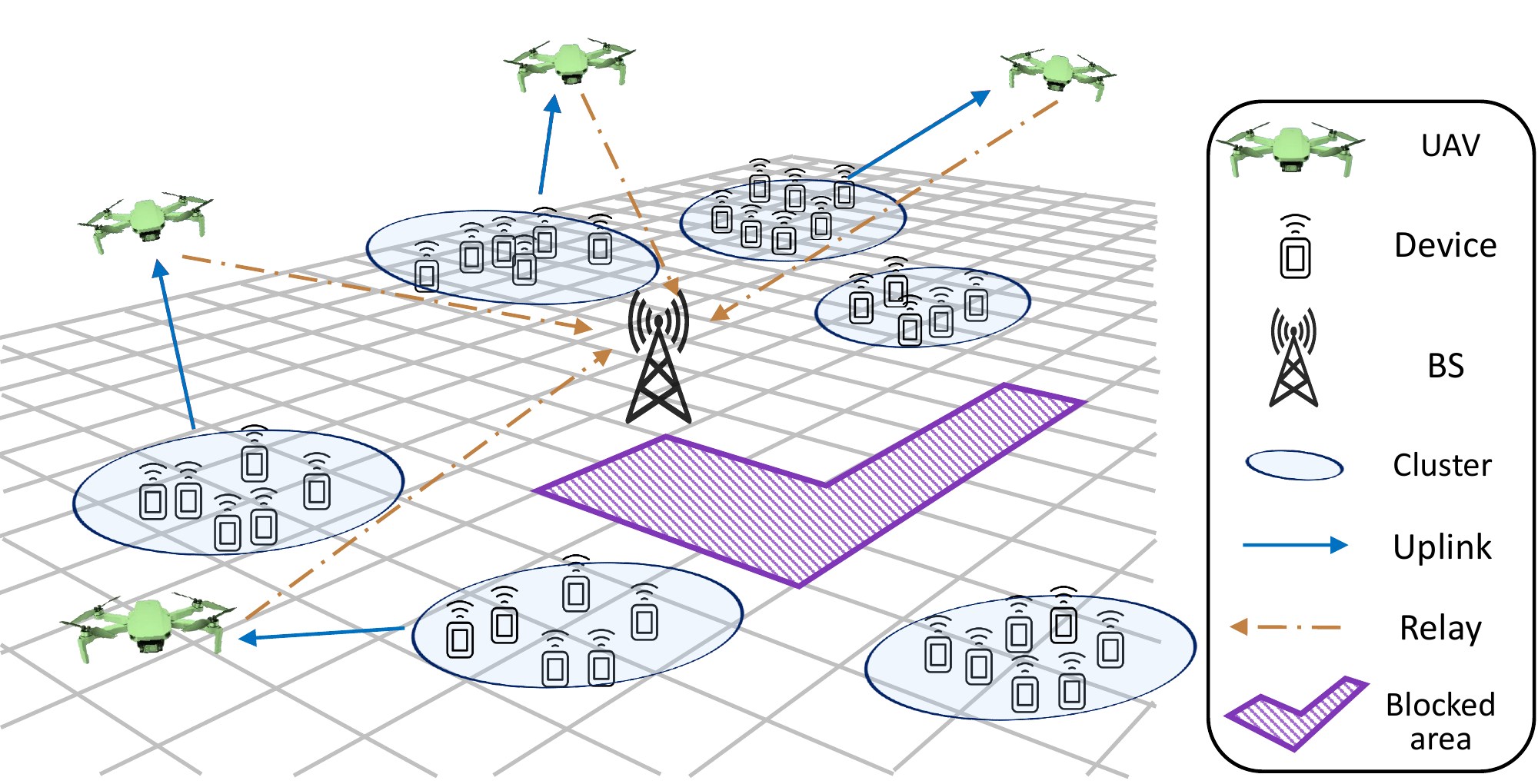} \vspace{1mm}
	\caption{The system model: A swarm of UAVs flies around to receive information from clusters of IoT devices and relay this information to the BS. The blocked area is a no-fly zone.} \vspace{0mm}
	\label{System_Model_Fig}
\end{figure}

As illustrated in Fig.~\ref{System_Model_Fig}, we consider the uplink of a static and densely grounded IoT scenario with one BS with height $h_{BS}$ located in the center of a grid world. The vertical/horizontal distance between two adjacent cells is $L_c$. The set of IoT devices $\mathcal{D}=\{1,2,\cdots,D\}$ are uniformly distributed and each has a coordinate $(x_d,y_d)$. There exists a UAV swarm set $\mathcal{U} =\{1,2,\cdots, U\}$ of $U$ fixed-velocity rotary-wing UAVs that serve the devices and relay the information to the BS. There is a restricted area in the grid world in which the UAVs are not allowed to fly over for security reasons. We refer to the coordinates of the restricted area with the set $\mathcal{R}$. %=\{(x_1,y_2),\cdots,(x_R,y_R)\}$. 
Each UAV has three main tasks at each time instant: schedule one or more devices %$k_d(t)$
for uplink, move in a certain direction (or hover) $w(t)$, and relay the received packets to the BS. To this end, if a device $d$ is served by a UAV, we refer to this as $k_u(t) = d$, where the scheduling vector of all UAVs is $\boldsymbol{K(t)} = \left[k_1(t), \cdots, k_U(t)\right]$. In addition, each UAV can navigate as follows
\begin{equation} \label{eqn:directions}
	l_u(t+1)=
	\begin{cases}
		l_u(t)+(0,L_c), & \quad w_u(t)=\text{north}, \\
		l_u(t)-(0,L_c), & \quad w_u(t)=\text{south}, \\
		l_u(t)+(L_c,0), & \quad w_u(t)=\text{east}, \\
		l_u(t)-(L_c,0), & \quad w_u(t)=\text{west}, \\
		l_u(t), & \quad \text{Hover}, \\
	\end{cases}
\end{equation}
where $l_u = (x_u,y_u)$ is the coordinate vector of the UAV and the position vector of all the UAVs is $L_U = \left[l_1, \cdots, l_U\right]$. The movement vector of all the UAVs is $\boldsymbol{W(t)} = [w_1(t), \cdots, w_u(t)]$. The UAVs perform their tasks in the form of a sequence of cycles. Each cycle consists of 4 stages: \textit{(i) navigation stage}, \textit{(ii) transmission stage}, \textit{(iii) relaying stage}, and \textit{(vi) AoI update stage}. We define the unit of time in the network as a frame, where a UAV performs a full cycle in one frame. Each frame has a duration of $T_{fr}$ seconds. As illustrated in Fig.~\ref{UAV_Cycle}, we assume that the navigation stage is performed simultaneously with the other stages, where in the half-duplex mode, the transmission stage is followed by the relaying stage, whereas in the full-duplex mode\footnote{To facilitate the problem formulation, we assume ideal full-duplex operation. In practice, there may exist a performance degradation due to self-interference and hardware impairments, even though the self-interference can be brought down below the noise floor~\cite{book_Hirley}.}, both stages are performed simultaneously.

\subsection{Clustering, Rate Calculation, and UAV Cycle}
Consider a set $\mathcal{C} =\{1,2,\cdots,C\}$ of $\mathcal{C}$ clusters, where each device $d$ is assigned to a cluster $c$. Each cluster has a fixed number of devices denoted $D_c$. Assuming that each UAV schedules $D_c$ devices for uplink in the time dedicated for transmission $t_T$ given a transmission data rate $R_T$, the number of devices in each cluster is upper bounded by
\begin{equation}
    D_c \leq \frac{R_T \: t_T}{M},\label{eqn:rate_1}
\end{equation}
where $M$ is the packet size, $t_T$ is expressed in terms of the navigation time $t_N$, i.e., $t_T = \frac{t_N}{2} = \frac{L_c}{2 \: v_u}$ in the half-duplex mode and $t_T = t_N = \frac{L_c}{v_u}$ in the full-duplex mode. Note that, the extension to variable data rates is straightforward and does not lie within the direct scope of this work. Finally, the bound for the number of devices in a cluster is formulated as
\begin{equation}
D_c \leq
\begin{cases}
        \frac{R_T \: L_c}{2 \: M \: v_u}, & \quad \text{half-duplex}, \\
	\frac{R_T \: L_c} {M \: v_u}, & \quad \text{full-duplex}. \\
\end{cases}
\label{eqn:rate_2}
\end{equation}
Each UAV $u$ can schedule and serve $D_c$ devices in a cluster $c$ at each frame. We refer to the chosen clusters by $c$ by a UAV $u$ at time instant $t$ as $k_u(t) = c$. The scheduling vector of all UAVs is $K_U(t) = \left[k_1(t), \cdots, k_U(t)\right]$. The UAV cycle is illustrated as follows:
\begin{figure}[t!]
    \centering    \includegraphics[width=0.85\columnwidth]{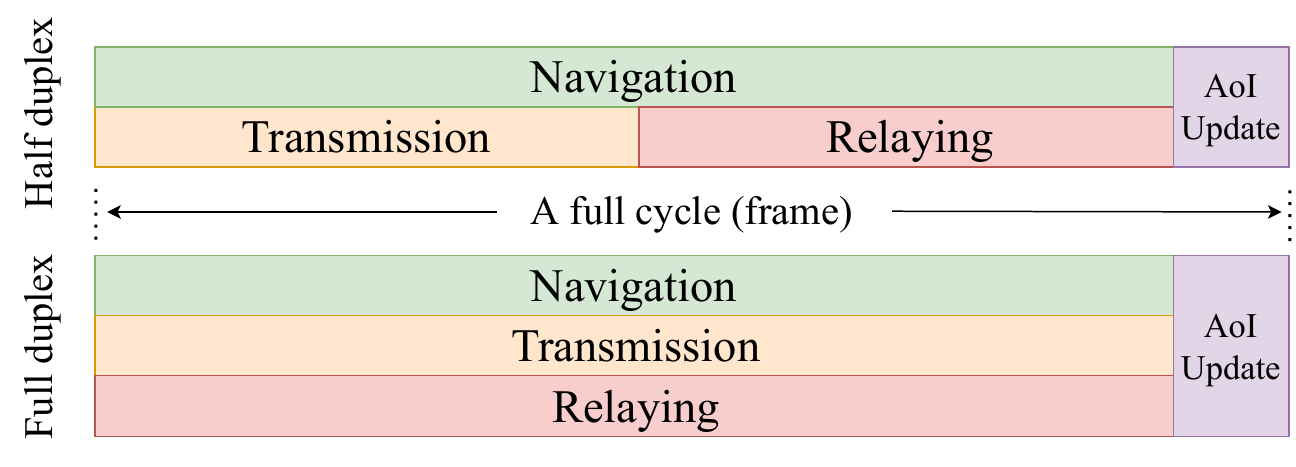} \vspace{1mm}
	\caption{A UAV navigation step from one grid point to another.} \vspace{0mm}
	\label{UAV_Cycle}
\end{figure}
\subsubsection{Navigation stage}
The position of each UAV $u$ at a time instant $t$ is determined by its height $h_u$ and the projection on the 2D plane $(x_u(t),y_u(t))$. Each UAV $u$ flies with a fixed velocity of $v_u$ from the center of a cell to the center of an adjacent cell. The duration of the navigation stage is formulated as follows

\begin{equation}
\label{t_Fly}
    t_N = \frac{L_c}{v_u}.
\end{equation}

\subsubsection{Transmission stage}
The UAVs communicate with the scheduled devices to receive their packets. We assume an LoS component between the UAV and the devices. When a UAV $u$ schedules an IoT device $d$ at time $t$, the channel gain between both is
\begin{equation}
\label{ch_gain_d}
    g_{du}(t) = \frac{\beta_0}{h_u^2+||L_{du}(t)||^2},
\end{equation}
where $\beta_0$ is the channel gain at a reference distance of $1$ m, $L_{du}$ is the distance between the device and the UAV. The required transmission power allocated for the device is
\begin{equation}
\label{Tx_P}
    P_d(t) = \! \frac{\left(2^{\frac{M}{B}}\!-1\right) \sigma^2}{g_{du}(t)} \!= 
    \left(2^{\frac{M}{B}}-1\right)\frac{\sigma^2}{\beta_0}\:\left( h_u^2+||L_{du}(t)||^2\right),
\end{equation}
where $M$ is the packet size, $B$ is the bandwidth and $\sigma^2$ is the noise power. The duration of the transmission stage is $t_T$ and it is half the duration of the navigation stage $t_T = \frac{t_N}{2}$ in the half-duplex mode and it equals the duration of the navigation stage $t_T = t_N$ in the full-duplex mode. The transmission power vector of the network is $\mathbf{P}(t) = \left(P_1(t),\cdots,P_D(t)\right)$.

\subsubsection{Relaying stage}
We assume an LoS component between the UAV and the BS. The channel gain between the UAV $u$ and the BS at time instant $t$ is given by \cite{deep_us}
\begin{equation}
\label{ch_gain_BS}
    g_{uBS}(t) = \frac{\beta_0}{|h_u-h_{BS}|^2+||l_u||^2},
\end{equation}
where the coordinate of the BS is assumed to be $(0,0)$. The duration of the relaying stage is $t_R$ and it is half the duration of the navigation stage $t_R = \frac{t_N}{2}$ in the half-duplex mode and it equals the duration of the navigation stage $t_R = t_N$ in the full-duplex mode.

\subsubsection{AoI update stage}
The AoI of device $d$ is calculated as the time difference between the current instant and the time instant of the last update received from device $d$. Thus, the AoI for device $d$ at time instant $t$ is updated according to the following update equation
\begin{equation}
\label{AoI_update}
	A_d(t) =
	\begin{cases}
		1, & \quad \text{if} \ s_d(t)=1, \\
		\text{min}\{A_{max},A_{d}(t-1) + 1\}, & \quad \text{otherwise}, 
	\end{cases}
\end{equation}
where $A_{max}$ is the maximum AoI threshold allowed in the network, which is chosen to be relatively high. The duration of the AoI update stage is neglected. The AoI vector of the network is $\mathbf{A}(t) = (A_1(t),\cdots,A_D(t))$.

\subsection{Problem Formulation}
The goal is to plan the trajectories of the UAVs and their scheduling policies to maximize the freshness of the information, i.e., minimize the average AoI of the IoT devices. Consider the weighting vector $\mathbf{\Theta} = (\theta_1,\cdots,\theta_D)$, where $\theta_d$ influences the importance of the AoI of device $d$. We can formulate the optimization problem as follow
\begin{subequations}\label{P1}
	\begin{alignat}{2}
	\mathbf{P1:}\qquad &\underset{\boldsymbol{W(t),K(t)}}{\min}       &\ \ \ & \frac{1}{T}\sum_{t=1}^T \Tr \left( \: {\mathbf{\Theta \: A}}\right) + \frac{\zeta}{D} \Tr {\mathbf{ \left(P\right)}}
,\label{P1:a}
	\ \\
	&\text{s.t.}   &      & l_u(t) \in \mathcal{X}, \label{P1:b}
	\end{alignat}
\end{subequations}
where $\Tr(\cdot)$ is the trace of a matrix, $\zeta$ is a transmission power penalty to the optimization problem to force the UAV to move closer to the devices with high AoI, and $\mathcal{X}$ is the set of all possible locations in the grid world. The constraint in \eqref{P1:a} ensures that the UAVs move inside the grid world, whereas the constraint \eqref{P1:b} ensures that the optimized parameter will not lead one of the UAVs inside the restricted area.

\section{The Proposed MARL Solutions}\label{sec:DRL} 
The aforementioned optimization problem in $\mathbf{P1}$ is a non-linear integer programming optimization problem whose order of complexity is very large, even for a small number of UAVs~\cite{9815722}. We propose novel centralized and decentralized MARL solutions to solve this complex optimization problem in a massive setup of IoT devices and a swarm of UAVs.

\subsection{The Markov Decision Process}
Assume that the swarm of UAVs are the agents that need to optimize their movement and scheduling decisions to minimize the objective function in ~\eqref{P1} in the environment denoted in the previous section. In deep reinforcement learning, Markov decision process (MDP) problems are usually interpreted in terms of the state, action, and reward, where the agent $u$ explores the environment at time $t$, observes a state $s(t)$, takes an action $a(t)$ that transits the agent to a new state $s(t+1)$. Then the agent receives a reward of $r(t)$. Therefore, for $\mathbf{P1}$, those are defined as follows:

\subsubsection{State}
it consists of the position of the UAV $l_u(t) = (x_u(t),y_u(t))$ and the AoI of the devices $\mathbf{A}(t)$. Therefore, the state vector of a UAV $u$ at time instant $t$ is $s_u(t) = \left[l_u,A(t)\right]$.

\subsubsection{Action space}
it consists of the movement direction of the UAV $w_u(t)$ and the chosen cluster by the UAV $k_u(t)$. Therefore, the action vector of a UAV $u$ at time $t$ is $a_u(t) = \left[w_u(t),k_u(t)\right]$.

\subsubsection{Reward}
it is formulated in terms of the AoI and the transmission power as defined in~\eqref{P1}. The immediate reward of a UAV $u$ at time $t$ is %\textcolor{red}{MS: could you write this as a trace function like P1 ?}
\begin{equation}
\label{Reward}
    r_u(t) = - \Tr \left( \: {\mathbf{\Theta \: A}}\right) - \frac{\zeta}{D_c} \Tr {\mathbf{ \left(P\right)}},
\end{equation}
where the goal is to maximize the accumulative reward
\begin{equation}
\label{reward_max}
\underset{\pi_u}{\max} \: \sum_{t=1}^{T} r_u(t),
\end{equation}
where $\pi_u$ is the policy followed by the UAV during the whole episode and $T$ is the episode length. %\textcolor{red}{MS: any discount factors here ?}

\subsection{MARL Solutions} \label{dqn_solution}
The Q-function $Q(s_u,a_u)$ evaluates how good an action is at each particular state~\cite{9815722}\footnote{Note that \cite{deep_us,9815722} provide interesting insights about age minimization as well as the air interface between UAVs and ground nodes. However, the solutions proposed therein lack scalability to massive networks served by a UAV swarm.}. It calculates the expected accumulative reward by taking an action $a_u$ at state $s_u$ and following the policy $\pi_u$. The optimal policy $\pi_u^*$ for each UAV $u$ is the policy that maximizes the Q-function, i.e., at each state, choose the action that has the highest expected accumulative reward 
\begin{equation}\label{optimal_policy}
\pi_u^* = \argmax_{a_u} \: Q(s_u,a_u),
\end{equation}
where the optimal policy of all the UAVs is $\pi^*=\left[\pi_1^*,\cdots,\pi_U^*\right]$. This problem can be solved iteratively, where
\begin{align}
& Q\left(s_u\left(t\right),a_u\left(t\right)\right) \leftarrow \:  Q\left(s_u\left(t\right),a_u\left(t\right)\right) + \nonumber\\
&\alpha \:  \left(r_u\left(t\right) +  \gamma \: \max_{a_u} Q\left(s_u\left(t+1\right),a_u\right)
-Q\left(s_u\left(t\right),a_u\left(t\right)\right)\right),
\end{align}
where $\alpha$ is the learning rate, and $\gamma$ is the discount factor that controls how much the network cares about the future rewards compared to the immediate rewards. Solving the Q-function iteratively requires the agents to visit all the possible states and experience all the possible actions, which is relatively complex in high dimension state and action spaces~\cite{8422586}. Instead, function approximators, such as neural networks, are used to overcome the dimensionality curse~\cite{DQNs}. We apply a deep Q-network (DQN) solution to find the optimal Q-function and the optimal policy.

In DQNs, a neural network is used as a function approximator to estimate the Q-function $Q(s_u,a_u|\theta_u)$, where $\theta$ is a vector containing the weights of the trained network. This network is called the current network. The Q-function is estimated by optimizing $\theta_u$ that minimizes the loss function
 \begin{align}
    \mathbf{L}(\theta_u(t)) = &(r_u(t)+\max_{a_u} Q(s_u(t+1),a_u|\theta_u(t-1))\\ \nonumber
    &-Q(s_u(t),a_u(t)|\theta_u(t)))^2.
\end{align}
 
In addition, the DQN adopts another neural network, which is called the target network to estimate the target Q-function values that will be used to update the Q-function~\cite{DQNs}. Herein, we propose four schemes to train the agents using the DQN algorithm as follows:
\subsubsection{\textbf{Centralized-RL (C-RL)}}
The training is done at the BS, which passes the optimal policy to the agents. This is a high-complexity scheme since the state and action spaces include all possible options for all the UAVs $s(t) = \left[L_U(t), A(t)\right]$ and $a(t) = \left[W(t), K_U(t)\right]$, respectively. Herein, the message exchange overhead is low, but it requires a very long training time to converge to the optimal policy.

\subsubsection{\textbf{Cooperative-MARL (Co-MARL)}}
The UAVs are considered to be individual agents and can train their own local DQNs. The training is run serially, where each UAV passes its selected action to the next UAV directly or through the BS \footnote{We assume that passing information about actions between UAVs is done within a negligibly short time, compared to the navigation step time $T_N$. Moreover, this work focuses on evaluating different methods for learning the trajectory of each UAV; therefore, to facilitate the analysis, we assume that the sharing of information between UAVs is performed through reliable error-free links.}. Each UAV utilizes the optimized policies of the other UAVs to find its optimal policy. The AoI is updated universally at the BS and distributed among all the UAVs at the end of each episode. This scheme is of low complexity since each UAV considers only its state and action spaces during training; however, it requires high message exchange between the UAVs which consumes time and decreases the spectral efficiency.

\subsubsection{\textbf{Partially cooperative-MARL (PCo-MARL)}}
Each UAV runs its training locally and they share its optimized actions with the BS without sending them to the other UAVs. The BS calculates the universal AoI and distributes it among all the UAVs. This scheme is simple and has low message exchange, but it does not have full knowledge of the true state and therefore, is unable to converge to the optimum policies in many cases.

\subsubsection{\textbf{Decentralized-MARL (D-MARL)}}
\begin{figure}[t!]
    \centering    \includegraphics[width=0.85\columnwidth]{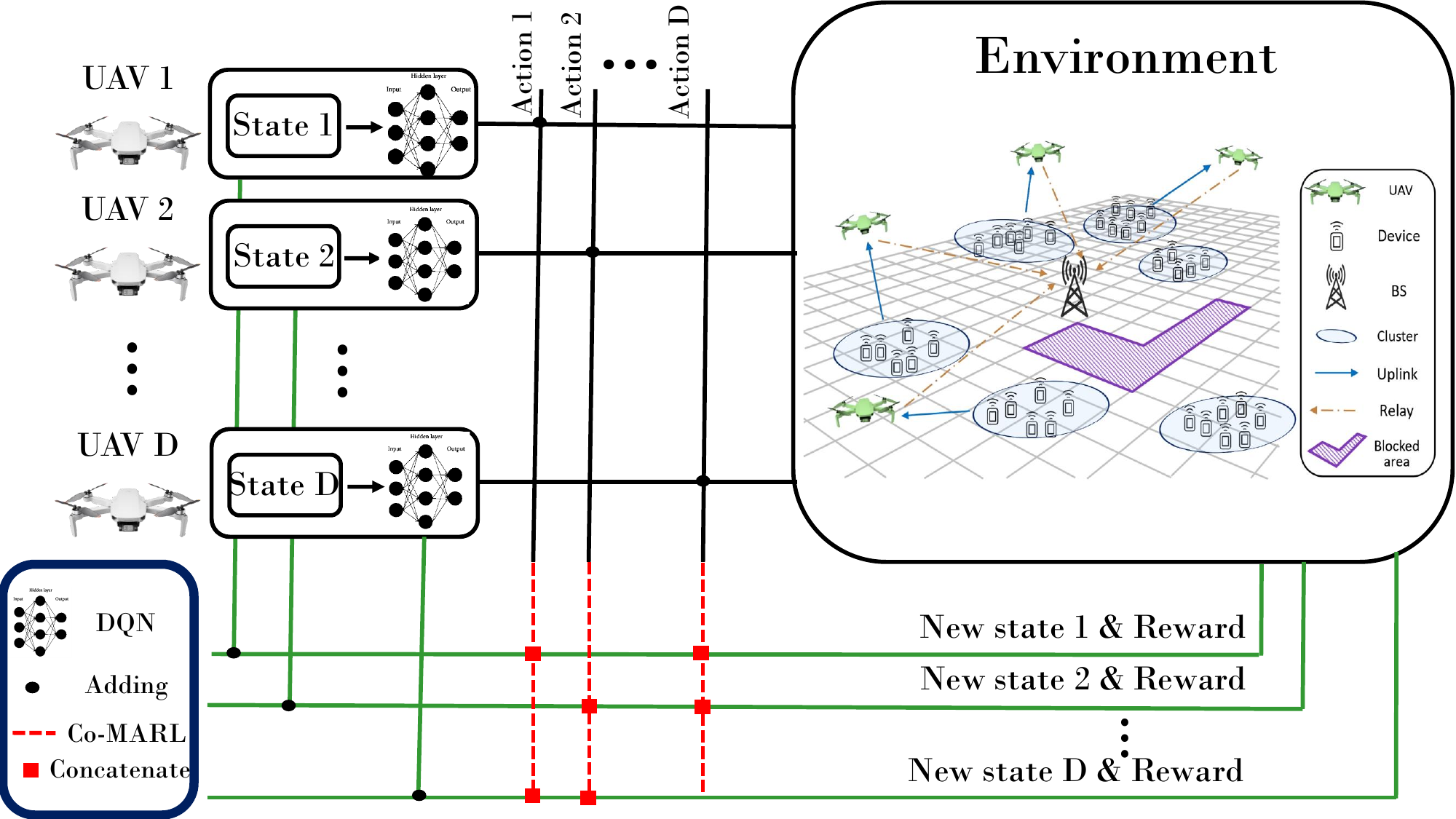} \vspace{1mm}
	\caption{A schematic for the proposed MARL algorithms} \vspace{0mm}
	\label{UAV_Algorithm}
\end{figure}

Each UAV runs its training locally without sharing any information with other UAVs or BS. This scheme is very simple and has high spectral efficiency, but its performance is poor as it possesses no information about the true states and actions of the other UAVs. Fig.~\ref{UAV_Algorithm} illustrates the proposed DQN solution and the difference between the Co-MARL and the D-MARL schemes.
%\textcolor{red}{how does it show that difference ?}
%%%%%%%%%%%%%%%%%%%%%%%%%%%%%%%%%%%%

\section{Numerical Results and Discussion}\label{results}

In this section, we show the simulation results of the proposed MARL solutions. We consider a $1100$ m $\times$ $1100$ m network with the parameters shown in Table~\ref{UAV_Parameters}. The training is performed using the Pytorch framework on a single NVIDIA Tesla V100 GPU. Due to the high number of UAVs, the trajectories would appear unclear and interfere with each other. Therefore, we are not providing the UAV trajectory plots here

\begin{table}[t!]
\centering
\caption{UAV model and DQN parameters.}
\label{UAV_Parameters}
\begin{tabular}{cc|cc}
\toprule
\textbf{Parameter}                                    & \textbf{Value} & \textbf{Parameter}                                    & \textbf{Value} \\ \midrule

$\beta_0$ & $30$ dB & $\sigma^2$ & $-100$ dBm \\
$B$ & $1$ MHz & $M$ & $5$ Mb \\
$h_u$ & $100$ m & $h_{BS}$ & $15$ m \\
$v_u$ & $25$ m/s & $L_c$ & $100$ m \\
$A_{max}$ & 30 & $\zeta$ & $5$ \\
$T$ & $60$ frames & $\theta_d$ & $\frac{1}{D}$ \\
$\alpha$ & $0.0001$ & $\gamma$ & $0.99$ \\
DQN layers & $(64,128,64)$ & Episodes & $100000$ \\
Loss & MSE & optimizer & Adam \\

\bottomrule
\end{tabular} 
\end{table}

\begin{figure*}[t]
    \centering
    \subfloat[Half-duplex \label{Rate_a}]{\includegraphics[width=0.4\textwidth]{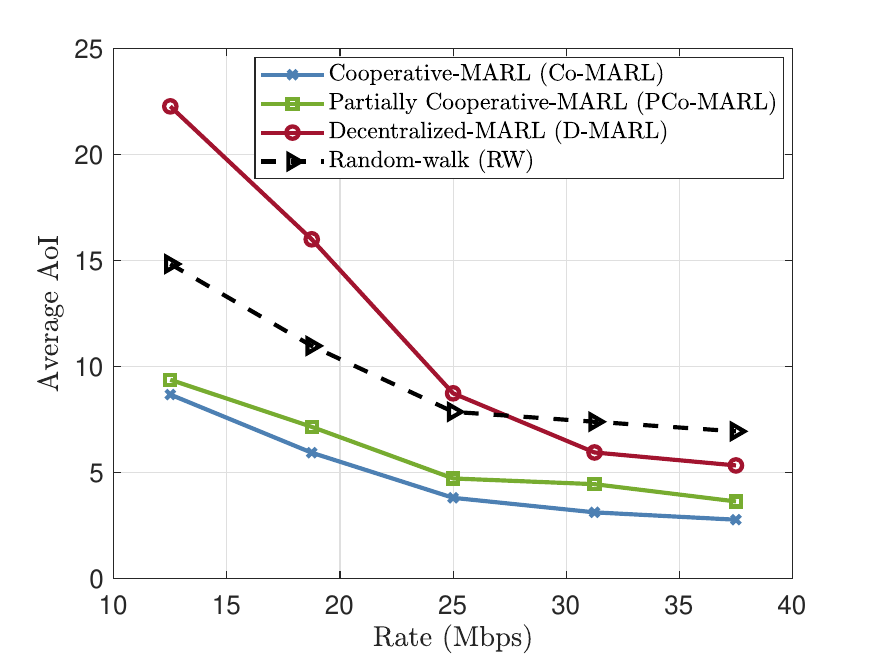}}
    \hskip -1.9ex
    \subfloat[Full-duplex\label{Rate_b}]{\includegraphics[width=0.4\textwidth,trim={0 0 0 0},clip]{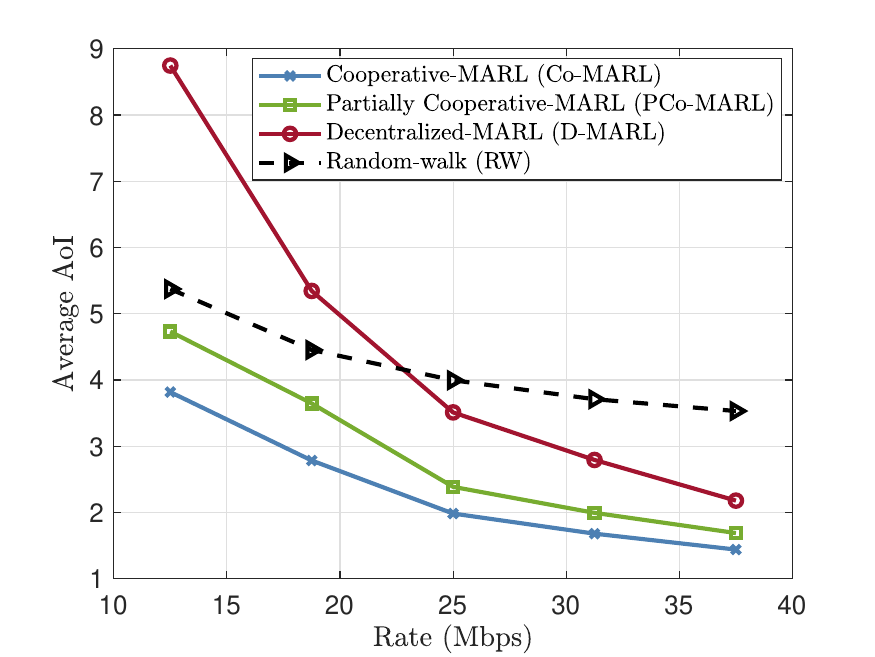}}
    \hskip -1.9ex
    \caption{The average AoI of $10$ UAVs serving $300$ IoT devices in the half-duplex and the full-duplex modes while changing the transmission rates.}
    \label{Rate} \vspace{-0mm}
\end{figure*}

Fig.~\ref{Rate} illustrates the average AoI in the half-duplex and the full-duplex modes for a swarm of $10$ UAVs that are serving $300$ IoT devices while varying the transmission rate. As illustrated in~\eqref{eqn:rate_2}, the transmission rate dictates the number of devices in each cluster, and hence, controls the number of clusters that has a direct influence on the AoI. The average AoI is reduced as the rate increases. In addition, the Co-MARL solution has the lowest AoI compared to the PCo-MARL and the D-MARL solutions. Overall, all the learning schemes outperform the RW baseline model. However, low transmission rates affect the D-MARL scheme forcing the UAVs to choose a particular action (a cluster and a movement direction) all the time as the action space becomes relatively large. Therefore, the decentralized solution only outperforms the RW in high transmission rates. In Fig.~\ref{Rate_a}, the D-MARL solution outperforms the RW for rates higher than $27.5$ Mbps in the case of the half-duplex mode, whereas in Fig.~\ref{Rate_b}, it renders better AoI for rates higher than $22.5$ Mbps. Moreover, at the same transmission rates, the full-duplex mode always outperforms the half-duplex mode, since the UAV is able to receive and transmit data simultaneously, which allows for a higher number of served devices per navigation step and hence, lower age.

\begin{figure}[t!]
    \centering\includegraphics[width=0.85\columnwidth]{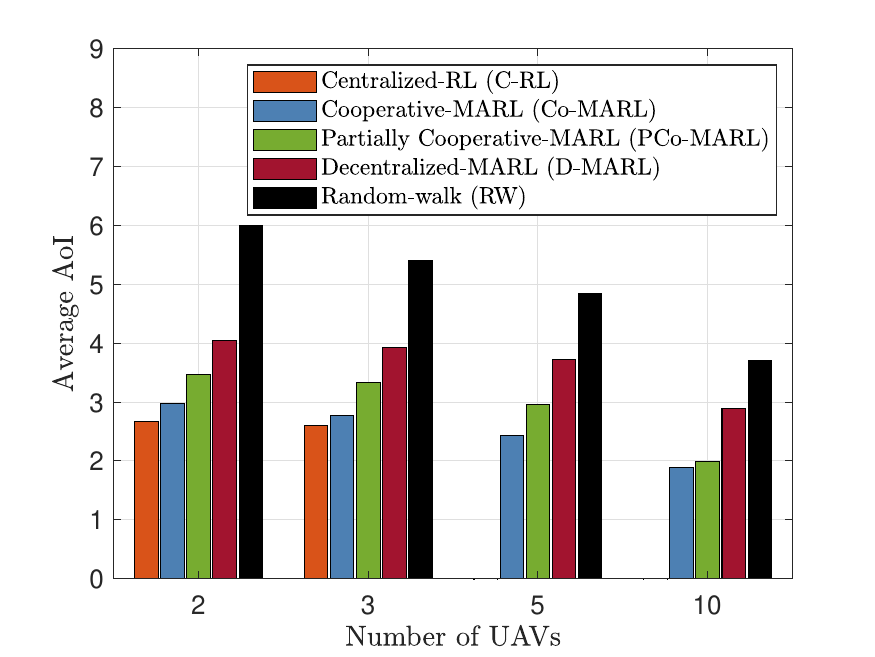} \vspace{0mm}
    \caption{The average AoI while changing the number of the UAVs that are serving $300$ IoT devices in the full-duplex mode using a transmission rate of $31.25$ Mbps.} 
    \vspace{0mm}
    \label{Number_UAVs}
    
\end{figure}

Fig.~\ref{Number_UAVs} shows the average AoI resulting from different numbers of UAVs that are serving $300$ devices in the full-duplex mode using a transmission rate of $31.25$ Mbps. All the MARL schemes are trained and compared using the same DQN and the same number of episodes. We can notice that as the number of UAVs increases, the AoI decreases for all schemes. All the proposed solutions outperform the RW baseline solution. The Co-MARL scheme has the lowest AoI, whereas the D-MARL scheme has the highest AoI compared to all the proposed schemes. The C-RL has the lowest AoI but it requires more episodes in training compared to the other schemes as its state and action spaces are relatively large. In addition, when the number of UAVs is larger than $3$ UAVs, the C-RL is unable to converge due to the dimensionality curse.

Table~\ref{TABLE_Time} depicts the complexity analysis of all the proposed schemes and the RW baseline scheme. We define the time complexity as the execution time of each algorithm, whereas we define the computational complexity ($\#$ computations) as the number of multiplications and additions required by each algorithm. The signaling overhead is the number of messages exchanged (other than the relayed data) between the BS and the UAVs. The RW has the lowest time complexity as it does not perform any computations. During the training, the C-RL requires double the number of episodes of the other schemes to converge. The Co-MARL scheme has the highest time complexity for an episode as each UAV must wait for the information shared by other UAVs before taking an action due to its high signaling overhead. The C-RL scheme has the lowest signaling overhead but it demands a large number of computations (10 times higher than MARL) and its time complexity is also higher. Meanwhile, all MARL schemes have lower computation demands.

\begin{table}[t!]
\centering
\caption{The complexity analysis per episode of $3$ UAVs serving $300$ devices using a transmission rate of $31.25$ Mbps.}
\label{TABLE_Time}
\begin{tabular}{c|c|c|c}
\toprule
\textbf{Parameter} & \textbf{Time complexity} & \textbf{$\#$ Computations}  & \textbf{Signaling} \\ \midrule
C-RL & $5.85$ ms & $7.74 \times 10^5$ & $3$ \\

Co-MARL & $5.10$ ms & $7.32 \times 10^4$ & $12$ \\

PCo-MARL & $4.53$ ms & $7.26 \times 10^4$ & $9$ \\

D-MARL & $4.47$ ms & $7.26 \times 10^4$ & $6$ \\

RW & $0.6$ ms & $0$ & $3$ \\

\bottomrule
\end{tabular}
\end{table}

Taking a deeper look into the results in Fig. \ref{Number_UAVs} and Table \ref{TABLE_Time}, one can observe that the deployment of a higher number of UAVs can reduce the average AoI of the network. For example, deploying $10$ UAVs instead of $3$ UAVs improves the average AoI to a value of $1.9$ instead of $2.6$ (i.e., about $30\%$ improvement). Moreover, $10$ UAVs are able to reach this low AoI value using MARL which requires much less UAV computation capabilities on board. The cost of a UAV with less processing capabilities on board could indeed be cheaper. An interesting conclusion here is that we are able to acquire more fresh information using a swarm of low-cost UAVs rather than deploying a small number of high-cost UAVs.

%%%%%%%%%%%%%%%%%%%%%%%%%%%%%%%%%%%%

\section{Conclusions}\label{conclusions} %\vspace{1mm}

This paper presented different learning approaches on how to collect fresh information from a massive IoT network using a UAV swarm. The main target of the proposed approaches was to maximize the overall information freshness of the network. We applied different MARL algorithms and compared them to the centralized RL solution. The results revealed that the, although the centralized solution is able to reach the best performance with a low number of UAVs, its time and computational complexity, are quite high. Moreover, it fails to converge for a high number of UAVs, unlike the MARL solutions which provide low-complexity solutions in all scenarios and are able to scale well to the UAV swarm setup. An interesting future direction is to consider the non-ideal transmission of information among UAVs and study its effect on the overall performance of MARL as noted in Section \ref{dqn_solution}.

\appendices 

\section*{Acknowledgments} \vspace{1mm}
This work is partially supported by Academy of Finland, 6G Flagship program (Grant no. 346208) and FIREMAN (Grant no. 326301), and the European Commission through the Horizon Europe project Hexa-X (Grant Agreement no. 101015956). 

\bibliographystyle{IEEEtran}
\bibliography{IEEEabrv,references}
\end{document}